\documentclass{article}

  \usepackage[dblblindworkshop, final]{neurips_2026}

\workshoptitle{The 6th Workshop on Mathematical Reasoning and AI}

\usepackage[utf8]{inputenc} 
\usepackage[T1]{fontenc}    
\usepackage{hyperref}       
\usepackage{url}            
\usepackage{booktabs}       
\usepackage{amsfonts}       
\usepackage{nicefrac}       
\usepackage{microtype}      
\usepackage{xcolor}         
\usepackage{amsmath}
\usepackage{graphicx}

\title{Order-Invariant Answers, Order-Sensitive Representations in Mathematical Reasoning}

\author{%
  Zhixu Silvia Tao\\
    Operations Research and Financial Engineering\\
  Princeton University\\
  Princeton, NJ 08540 
}

\begin{document}

\maketitle

\begin{abstract}
Reordering a set of mathematical rules without changing its meaning should preserve the correct answer, but must a model's internal representations stay invariant too? We investigate this question using synthetic multi-step function-composition problems, each presented under multiple rule orderings with the same correct answer. We measure accuracy and permutation signal-to-noise ratio (SNR), which quantifies how distinctly ordering patterns are represented relative to variation across problem instances. Across 16 language models ranging from 1B to 8B parameters, we find a pattern: models that solve reordered problems more accurately represent different rule orderings more distinctly. Layer-averaged permutation SNR is positively rank-correlated with accuracy in every synthetic setting we evaluate, with Spearman correlations reaching 0.86. These findings highlight a distinction between answer invariance and representation invariance: successful mathematical rule composition can accompany distinct internal representations between equivalent rule orderings. This motivates distinguishing answer invariance from representation invariance, and offers a representational perspective on mathematical reasoning beyond answer accuracy alone.
\end{abstract}

\section{Introduction}
Mathematical reasoning requires identifying the logical
relationships among definitions and premises, and following those relationships to reach a correct conclusion. When a problem is reformulated without changing its meaning, these relationships and the correct conclusion remain unchanged. Reordering a set of premises is one such transformation: it changes their presentation, but not what follows from them.
The task is therefore \emph{answer-invariant}: different
orderings require the same conclusion. One might expect
models that reason accurately across these orderings to
also form similar internal representations, treating
presentation order as irrelevant. But does accurate
reasoning on an answer-invariant task accompany
\emph{representation invariance}, or can models distinguish
different orderings internally while still reaching the
correct answer?

We investigate this question through synthetic multi-step
function-composition problems. Each problem lists numerical mappings such as $F(3)=7$ and $G(7)=-2$ alongside distractor mappings. Given the input $3$, the model must follow the relevant chain to obtain $-2$. Shuffling the rule lines changes where the mappings appear but still preserves the logical dependencies needed to solve the problem. The task therefore guarantees answer invariance, allowing us to examine how models represent different orderings of the same mathematical content.

Prior work shows that premise order can affect mathematical and deductive reasoning accuracy \citep{chen2024premise}, and investigates augmentation to improve reasoning across equivalent orderings \citep{he2025order}. We offer one more perspective from representational space analysis. One might expect models that solve reordered problems accurately to form a common internal representation that ignores rule order. We observe the opposite pattern: more accurate models represent different orderings
more distinctly.

To quantify this distinction, we define permutation
signal-to-noise ratio (SNR), a representation space statistic measuring how distinctly ordering patterns are represented across problem instances. It compares the separation between average representations for different orderings with variation across problems presented in the same ordering. Higher SNR indicates greater relative separation of ordering-specific representations.

Across 16 language models with 1B--8B parameters,
layer-averaged permutation SNR is positively rank-correlated with accuracy in all three evaluated synthetic settings, with Spearman correlations reaching $0.86$. Thus, models that solve reordered problems more accurately
represent different rule orderings more distinctly.
Accurate reasoning across orderings therefore coexists
with internal representations that distinguish those orderings, even though they share the same correct answer.

\paragraph{Contributions}
Our contributions are threefold. We (i) construct a controlled rule-shuffling task that varies presentation order while holding problem content and the correct answer fixed; (ii) define permutation SNR to quantify permutation-pattern separability in hidden representations; (iii) show across 16 models and three synthetic settings that models with higher accuracy distinguish rule-ordering patterns more clearly in their representations. Together, these provide a representational perspective on mathematical reasoning that complements final-answer accuracy.

\section{Task and Representation Measure}
\label{sec:task}

\subsection{Rule-Shuffling Task}
\label{sec:rule-task}

We consider mathematical problems that require following a
sequence of numerical function mappings. For example, suppose a problem provides
\[
F(3)=7,\quad G(7)=-2,\quad F(5)=9,\quad G(9)=4,
\]
and asks for $G(F(3))$. The model must identify the relevant chain, $3 \xrightarrow{F} 7 \xrightarrow{G} -2$,
while ignoring the mappings involving input $5$.
Listing these same rules in a different order, such as
\[
G(9)=4,\quad G(7)=-2,\quad F(5)=9,\quad F(3)=7,
\]
preserves the relevant chain and the correct answer, $-2$. More generally, each problem contains a relevant chain
\[
f_1(a_0)=a_1,\quad \ldots,\quad f_D(a_{D-1})=a_D,
\]
where $D$ is the composition depth. Given $a_0$, the model
must determine $a_D$. The rule set also contains $m$
distractor chains using the same function symbols with
different numerical values. Values at each stage are
distinct across chains, ensuring a unique path from the
queried input to the answer.

Each problem therefore contains $(m+1)D$ rule lines.
We present the same problem under different permutations
of these lines, changing their positions while preserving
their content and logical dependencies. This construction
holds the correct answer fixed while allowing us to examine
how rule order affects model accuracy and representations.

\subsection{Permutation Signal-to-Noise Ratio}
\label{sec:snr}

For each task setting, we consider $T$ problem instances,
each presented under $P$ shared permutation patterns.
A problem instance specifies the numerical rules and the
query; a permutation pattern specifies the order in which
those rules are presented. For example, the four rules in \S\ref{sec:rule-task}, together with the query for input $3$, form one problem instance. Label its rules as
\[
r_1:F(3)=7,\quad r_2:G(7)=-2,\quad
r_3:F(5)=9,\quad r_4:G(9)=4.
\]
Applying the permutation pattern $(4,2,3,1)$ presents
this instance as
\[
G(9)=4,\quad G(7)=-2,\quad F(5)=9,\quad F(3)=7,
\]
while leaving the query and correct answer unchanged.
A different problem instance uses different numerical
values. Applying the same pattern $(4,2,3,1)$ to that
instance again places its fourth rule first, followed
by its second, third, and first rules.

We group representations by permutation pattern:
each of the $P$ groups contains the representations
of all $T$ problem instances presented under the same shuffling pattern. Permutation SNR measures the separation between these group centers relative to the variation within each group.

For a fixed model, let $h_{t,p}^{(\ell)}$ denote the
representation of problem $t$ under permutation $p$
at layer $\ell$. We obtain this vector by mean-pooling
hidden states over nonpadding prompt tokens before
answer generation. The center of permutation cluster
$p$ and the overall center are
\[
\mu_p^{(\ell)}
=\frac{1}{T}\sum_{t=1}^{T}h_{t,p}^{(\ell)},
\qquad
\mu^{(\ell)}
=\frac{1}{P}\sum_{p=1}^{P}\mu_p^{(\ell)}.
\]
We define
\[
\mathrm{SNR}_{\mathrm{perm}}^{(\ell)}
=
\frac{
  \underbrace{
    \frac{1}{P}\sum_{p=1}^{P}
    \left\|\mu_p^{(\ell)}-\mu^{(\ell)}\right\|_2^2
  }_{\text{between-cluster variation}}
}{
  \underbrace{
    \frac{1}{PT}\sum_{p=1}^{P}\sum_{t=1}^{T}
    \left\|h_{t,p}^{(\ell)}-\mu_p^{(\ell)}\right\|_2^2
  }_{\text{within-cluster variation}}
  +\epsilon
},
\]
where $\epsilon>0$ ensures numerical stability. Higher permutation SNR indicates more separated
ordering-specific cluster centers relative to variation among problems sharing an ordering. We use this relative separation to quantify
\emph{permutation-pattern separability}.
Finally, we average over $L$ hidden layers to obtain one value per model and task setting:
\[
\overline{\mathrm{SNR}}_{\mathrm{perm}}
=
\frac{1}{L}\sum_{\ell}
\mathrm{SNR}_{\mathrm{perm}}^{(\ell)}.
\]
\section{Experiments and Results}
\label{sec:experiments}
\subsection{Experimental Setup}
\label{sec:setup}

\paragraph{Models and tasks.}
We evaluate 16 instruction-tuned language models spanning
1B--8B parameters from the Qwen2.5, Llama, Mistral, Gemma,
and NVIDIA Ace families, including general-purpose, code,
and math variants (see Appendix~\ref{app:models} for full list). We consider three synthetic settings,
$(D,m)\in\{(2,4),(2,7),(3,4)\}$,
varying composition depth and distractor count.
Each setting contains $T=100$ problem instances, each
presented under the same $P=50$ permutation patterns,
yielding 5,000 prompts per model. Details of problem generation and rule-order sampling
are provided in Appendix~\ref{app:rule_sampling}.

\paragraph{Accuracy.}
We generate one sampled response per prompt and score whether
the extracted numerical answer matches the ground truth.
Accuracy is averaged over all problem instances and orderings.
Full prompts, decoding settings, and answer-extraction
details are provided in Appendix~\ref{app:setup}.

\paragraph{Representation measure.}
We compute permutation SNR from mean-pooled prompt representations
before answer generation, following \S\ref{sec:snr}.
We average the layerwise SNR values over the embedding output
and all transformer-layer outputs to obtain one value per
model and setting. We then compute the Spearman correlation
between accuracy and layer-averaged permutation SNR across
the same 16 models, separately within each setting.

\subsection{Results}
\label{sec:results}
\begin{figure}[t]
    \centering
    \includegraphics[width=\linewidth]
    {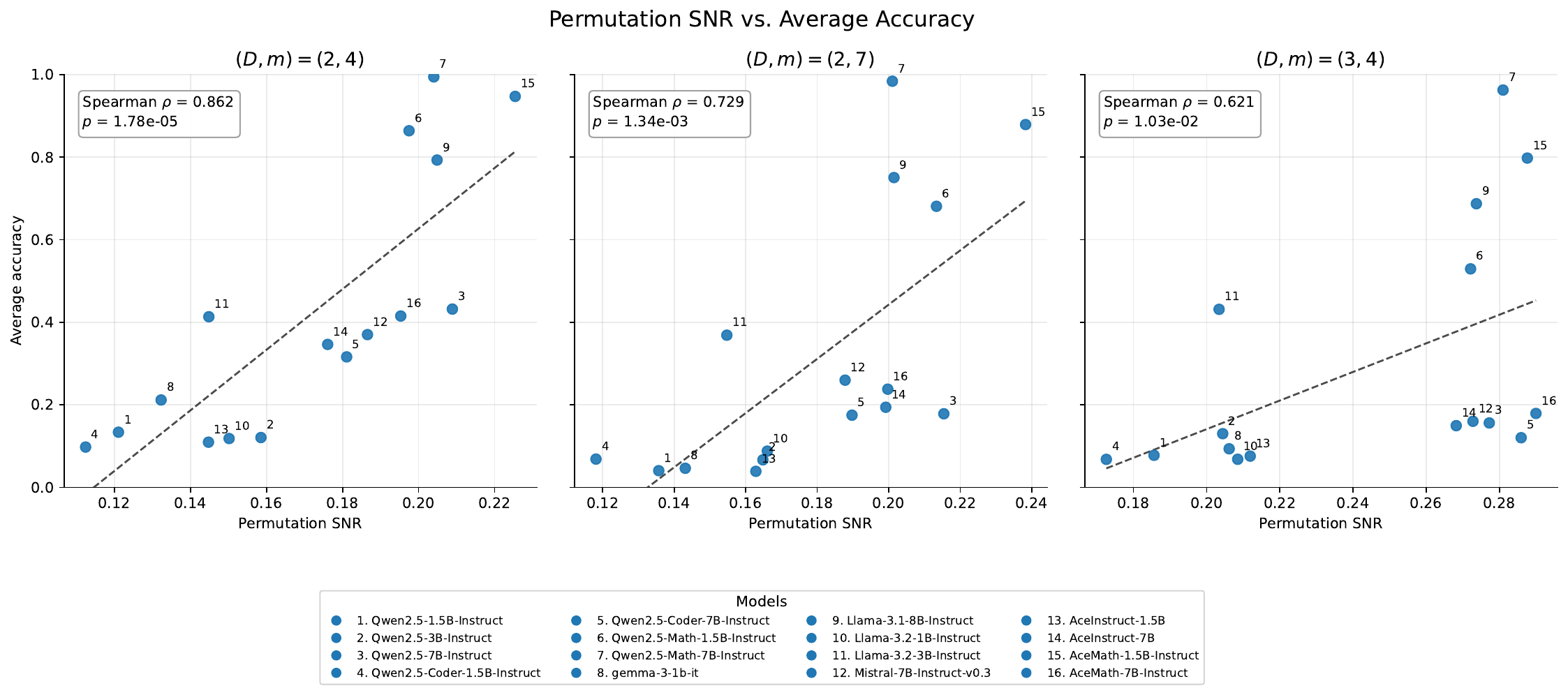}
    \caption{
        Layer-averaged permutation SNR versus average accuracy. Each panel compares the same 16 models under a different synthetic setting, with model identifiers shared across panels. Spearman correlations are computed within each setting. Dashed lines are linear fits shown as visual guides.
    }
    \label{fig:snr-vs-acc}
\end{figure}

\noindent\textbf{Higher reasoning accuracy is associated with greater permutation-pattern separability, although the association is less robust in the depth-three setting.}
Figure~\ref{fig:snr-vs-acc} shows the relationship between layer-averaged permutation SNR and average accuracy. Each panel compares the same 16 models within one setting. Spearman correlations are $\rho=0.86$, $0.73$, and $0.62$
for $(D,m)=(2,4)$, $(2,7)$, and $(3,4)$, respectively. Thus, models with higher permutation SNR tend to rank higher in accuracy across the evaluated composition depths
and distractor counts. 

We additionally examine this correlation after controlling for model size in Appendix~\ref{app:size-control}. The association remains statistically significant in both depth-two settings but weakens in the depth-three setting ($\rho_{\mathrm{partial}}=0.469$, $p=0.078$). The adjusted evidence for $(D,m)=(3,4)$ is therefore positive but inconclusive. One possible explanation is that increasing composition depth also increases task difficulty, concentrating
most models at relatively low accuracy and making the cross-model rank correlation more sensitive to small performance differences.

The direction of this relationship is counterintuitive. Changing the order of a problem's rules does not change its correct answer. One might therefore expect models
that solve these reordered problems accurately to
represent them similarly, treating rule order as
irrelevant. Instead, higher accuracy accompanies greater permutation-pattern separability: representations grouped by the same ordering pattern are more clearly separated relative to the variation within each group. Thus, models that solve reordered problems more successfully encode ordering patterns more distinctly.


This finding highlights the distinction between
\emph{answer invariance} and \emph{representation invariance}. Different rule orderings require the same answer, but a model can represent those orderings differently and still answer correctly. Greater accuracy across equivalent orderings is therefore associated with more distinct representations of ordering patterns. The positive
unadjusted correlations suggest that this association recurs within the evaluated task family, but the size-adjusted analysis indicates that its robustness decreases as compositional difficulty increases. The present results should consequently be interpreted as evidence of a cross-model association rather than evidence that permutation-pattern separability causally improves
reasoning.

\section{Conclusion and Limitations}
\label{sec:conclusion}


Across three synthetic function-composition settings, models with higher accuracy exhibit greater permutation-pattern separability. This finding highlights the distinction between answer invariance and representation invariance: solving equivalent rule orderings accurately can accompany distinct internal representations of those orderings. However, the robustness of this association varies across settings. In the depth-three setting, the correlation weakens and becomes inconclusive after controlling for model size, leaving open whether the association persists as compositional difficulty increases.


Our evidence is correlational and limited to synthetic tasks and models with partly shared ancestry. It therefore does not establish that encoding order improves reasoning or explain why more accurate models distinguish ordering patterns more clearly. Moreover, because we do not include a no-distractor baseline or fully cross composition depth with distractor count, we cannot isolate the effects of distractors, dependency depth, and prompt length. Permutation SNR measures the separability of shared ordering patterns, but may also capture superficial positional structure rather than representations of logical dependencies. Future work should evaluate matched depth--distractor conditions, including $m=0$, test the relationship within model families, intervene on ordering-related representations, and examine whether the findings extend to broader mathematical tasks.
\begin{ack}
The author thanks Noah Flynn, Matthew Trager and Henry C. Conklin for helpful discussions and feedback on this work, as well as the anonymous reviewers for their constructive comments. This work used computational resources provided by Princeton University.
\end{ack}

\bibliographystyle{plainnat}
\bibliography{ref}


\appendix
\section{Related Work}
\label{app:related-work}

\paragraph{Rule-based and symbolic reasoning.}
Language models have been evaluated on their ability to
derive conclusions from supplied premises and explicit rules \citep{saparov2023testing,hoppe2025investigating}.
Benchmarks such as ProofWriter, FOLIO, and LogicBench examine logical inference across different rule systems and problem formats \citep{tafjord2020proofwriter,han2022folio,parmar2024logicbench}. Synthetic datasets support controlled investigations of proof complexity, rule structure, and generalization \citep{saparov2022language,saparov2023testing,morishita2023learning}.
Our function-composition task follows this approach,
using numerical mappings and distractor chains to study
elementary mathematical reasoning. We vary the presentation
order while preserving the rules and correct answer,
and examine how representations of ordering patterns relate
to accuracy.

\paragraph{Order sensitivity in reasoning.}
Reordering a fixed set of premises preserves its logical
consequences, yet can affect language-model performance.
\citet{chen2024premise} demonstrate premise-order effects
in deductive and mathematical reasoning, while
\citet{he2025order} use augmentation over logically
equivalent orderings to improve reasoning performance.
Related work studies order independence through
Set-Based Prompting \citep{mcilroy2024order} and
answer-option order effects in analogical reasoning
\citep{lewis2024evaluating}.
Irrelevant information in mathematical problems can also
reduce accuracy \citep{shi2023large,mirzadeh2024gsm}.
We complement these studies by examining the relationship
between performance under reordering and the separation
of ordering patterns in hidden representations.

\paragraph{Representation geometry and reasoning.}
Prior work examines geometric structure associated with
truth and semantic concepts in language-model representations \citep{marks2023geometry,park2023linear}, as well as changes across layers and training stages
\citep{valeriani2023geometry,skean2025layer,li2025tracing}. More directly related to mathematical reasoning,
\citet{ye2025physics} use controlled synthetic problems
to investigate models' internal reasoning processes.
These studies motivate examining representations alongside
model outputs. Our analysis focuses on a specific property: how distinctly models represent different rule-ordering patterns across numerical problem instances.
Permutation SNR compares the separation of ordering-specific group means with the variation within those groups. We relate this statistic to accuracy across models, examining how successful reasoning on an answer-invariant task coexists with order-dependent representations.

\section{Experimental Setup Details}
\label{app:setup}
\subsection{Models}
\label{app:models}

We evaluate the following 16 instruction-tuned language models,
spanning approximately 1B--8B parameters:

\begin{itemize}
    \item \texttt{Qwen/Qwen2.5-1.5B-Instruct}
    \item \texttt{Qwen/Qwen2.5-3B-Instruct}
    \item \texttt{Qwen/Qwen2.5-7B-Instruct}
    \item \texttt{Qwen/Qwen2.5-Coder-1.5B-Instruct}
    \item \texttt{Qwen/Qwen2.5-Coder-7B-Instruct}
    \item \texttt{Qwen/Qwen2.5-Math-1.5B-Instruct}
    \item \texttt{Qwen/Qwen2.5-Math-7B-Instruct}
    \item \texttt{meta-llama/Llama-3.2-1B-Instruct}
    \item \texttt{meta-llama/Llama-3.2-3B-Instruct}
    \item \texttt{meta-llama/Llama-3.1-8B-Instruct}
    \item \texttt{mistralai/Mistral-7B-Instruct-v0.3}
    \item \texttt{google/gemma-3-1b-it}
    \item \texttt{nvidia/AceInstruct-1.5B}
    \item \texttt{nvidia/AceInstruct-7B}
    \item \texttt{nvidia/AceMath-1.5B-Instruct}
    \item \texttt{nvidia/AceMath-7B-Instruct}
\end{itemize}

\subsection{Problem Generation and Rule Ordering}
\label{app:rule_sampling}

We generate 100 problem instances for each setting
$(D,m)\in\{(2,4),(2,7),(3,4)\}$, where $D$ is the
composition depth and $m$ the number of distractor chains. 

For each instance, we construct $m+1$ chains, each containing $D+1$ numerical values connected by $D$ function mappings. At each position along the chains, we sample $m+1$ distinct integers uniformly without replacement from $\{-100,\ldots,100\}$ and assign one to each chain. Sampling is independent across positions. Distinct values at each position ensure that the queried input determines a unique chain and answer. The first chain supplies the query and ground-truth answer; the remaining chains serve as distractors.

Before shuffling, we arrange the rules chain by chain.
The rules needed to answer the query come first, in
function-application order. The distractor chains follow,
each listed in the same manner. For example, when the
queried input is $3$, this reference ordering is
\[
\underbrace{F(3)=7,\quad G(7)=-2}_{\text{queried chain}},
\qquad
\underbrace{F(5)=9,\quad G(9)=4}_{\text{distractor chain}}.
\]
Thus, the relevant rules appear first and consecutively,
with each mapping preceding the mapping that uses its output.

We assign indices to the $(m+1)D$ rule lines in this reference order. We then sample 49 distinct non-reference permutations of these indices uniformly, rejecting duplicates. Together with the reference ordering, these form a bank of 50 permutation patterns.

The bank is sampled once per setting and applied to every
problem instance. Each permutation index therefore specifies the same rearrangement of corresponding rule positions across problem instances. All models receive the same 5,000 prompts within a setting, and reported accuracy is averaged over all instances and all 50 orderings.

\subsection{Prompt Format}
\label{app:prompts}

All problems use the following system prompt:
\begin{quote}
\small\ttfamily\raggedright
You are solving rule-based reasoning problems. Give your final answer on its own line in exactly this format:\par
Answer: X\par
where X is the numeric value only, with no units or extra text after it.
\end{quote}

For composition depth $D=2$, the user prompt is:
\begin{quote}
\small\ttfamily\raggedright
Question: Below is a set of rules defining
F(x) = y, G(y) = z:\par
[rule lines in the selected ordering]\par
Given x = [input value], based on these rules, what is the
value of z?
\end{quote}

For $D=3$, the function sequence becomes
\texttt{F(x) = y, G(y) = z, H(z) = u}, and the query asks
for \texttt{u}. The bracketed fields are placeholders
replaced by the concrete rule lines and queried input.
Across permutations of a problem, only the order of the
numerical rule lines changes; the function sequence,
queried input, and final variable remain fixed.

We format the system and user messages using the model's
tokenizer chat template, with
\texttt{add\_generation\_prompt=True}.
If no chat template is available, we concatenate the
system and user texts with a newline.
The instructions specify the final-answer format but
neither require nor prohibit intermediate reasoning.

\subsection{Representation Extraction}
\label{app:representations}

We extract hidden states in a separate forward pass using
Hugging Face Transformers with the model in evaluation mode. Extraction uses the same prompt-formatting procedure as answer generation, with input truncation disabled. The representations contain only prompt tokens; no answer tokens are generated.

At each extracted hidden-state level, we mean-pool over
all nonpadding prompt tokens, including the system
instructions, numerical rules, and query. We exclude
padding positions using the attention mask and convert
the pooled vectors to 32-bit floating point before
computing permutation SNR.

We use $\epsilon=10^{-8}$ for numerical stability.
For each model and synthetic setting, we compute the
arithmetic mean of the layerwise SNR values, including
the embedding output (layer 0) and all returned
transformer-layer outputs. This produces one
layer-averaged permutation SNR value per model and setting.

\section{Controlling for Model Size}
\label{app:size-control}

Model size may contribute to both reasoning accuracy and
representation structure. To assess whether the observed
accuracy--SNR association persists after adjusting for size, we compute partial Spearman correlations separately within each synthetic setting, using the same 16 models.

For each model, we obtain its nominal parameter count from
the checkpoint identifier. We rank-transform accuracy,
layer-averaged permutation SNR, and log parameter count,
assigning average ranks to ties. We then regress ranked
accuracy and ranked SNR separately on ranked parameter
count, including an intercept, and compute the Pearson
correlation between the two sets of residuals. This yields the partial Spearman correlation controlling
for model size. Because the size variable is ranked,
using parameter count or its logarithm gives the same result.

\begin{table}[htp!]
    \centering
    \small
    \begin{tabular}{lccc}
        \toprule
        Setting $(D,m)$
        & Spearman $\rho$
        & Partial Spearman $\rho$
        & Partial $p$ \\
        \midrule
        $(2,4)$ & $0.862$ & $0.815$ & $2.13\times10^{-4}$ \\
        $(2,7)$ & $0.729$ & $0.647$ & $9.11\times10^{-3}$ \\
        $(3,4)$ & $0.621$ & $0.469$ & $0.078$ \\
        \bottomrule
    \end{tabular}
    \caption{
        Accuracy--SNR correlations before and after adjustment
        for model size. Each row uses the same 16 models.
        The final column reports approximate two-sided
        $p$-values for the partial correlations.
    }
    \label{tab:size-control}
\end{table}

The adjusted correlations remain positive in all three
settings, although their magnitudes decrease
(Table~\ref{tab:size-control}). The association remains
statistically significant at the $0.05$ level in both
two-step settings. In the three-step setting, the adjusted
correlation is positive but does not reach this threshold
($\rho=0.469$, $p=0.078$), leaving the evidence after
size adjustment inconclusive.

We calculate approximate two-sided $p$-values using
\[
t=\rho_{\mathrm{partial}}
\sqrt{\frac{n-3}{1-\rho_{\mathrm{partial}}^2}},
\]
with $n-3=13$ degrees of freedom.
These analyses adjust for ranked parameter count but
do not account for shared model ancestry, training data,
or specialization. They therefore do not establish
independence from other model characteristics or a
causal relationship between permutation SNR and accuracy.

\end{document}